\documentclass[10pt,twocolumn,letterpaper]{article}

\usepackage{iccv}
\usepackage{times}
\usepackage{epsfig}
\usepackage{graphicx}
\usepackage{amsmath, mathtools}
\usepackage{amssymb}

\usepackage{multirow}
\usepackage{makecell}
\usepackage{color}
\usepackage[dvipsnames]{xcolor}
\usepackage[font=normal]{subfig}
\usepackage[export]{adjustbox}
\usepackage{setspace}
\usepackage{booktabs}
\usepackage{array}

\usepackage[pagebackref=true,breaklinks=true,colorlinks,bookmarks=false]{hyperref}

\iccvfinalcopy 

\def\iccvPaperID{28} 

\DeclareSymbolFont{extrasymbols}{OMS}{cmsy}{m}{n}
\DeclareMathDelimiter{\lVert}
  {\mathopen}{extrasymbols}{"6B}{largesymbols}{"0D}
\DeclareMathDelimiter{\rVert}
  {\mathclose}{extrasymbols}{"6B}{largesymbols}{"0D}

\ificcvfinal\fi

\begin{document}

\title{AIM 2019 Challenge on Video Temporal Super-Resolution: Methods and Results}

\author{
    Seungjun Nah \and Sanghyun Son \and Radu Timofte \and Kyoung Mu Lee
    \and Li Siyao \and Ze Pan \and Xiangyu Xu \and Wenxiu Sun
    \and Myungsub Choi \and Heewon Kim \and Bohyung Han \and Ning Xu 
    \and Bumjun Park \and Songhyun Yu \and Sangmin Kim \and Jechang Jeong
    \and Wang Shen \and Wenbo Bao \and Guangtao Zhai \and Li Chen \and Zhiyong Gao
    \and Guannan Chen \and Yunhua Lu \and Ran Duan \and Tong Liu \and Lijie Zhang
    \and Woonsung Park \and Munchurl Kim
    \and George Pisha \and Eyal Naor \and Lior Aloni
}
    

\maketitle
\ificcvfinal\thispagestyle{empty}\fi

\begin{abstract}
   Videos contain various types and strengths of motions that may look unnaturally discontinuous in time when the recorded frame rate is low.
   This paper reviews the first AIM challenge on video temporal super-resolution (frame interpolation) with a focus on the proposed solutions and results.
   From low-frame-rate (15 fps) video sequences, the challenge participants are asked to submit higher-frame-rate (60 fps) video sequences by estimating temporally intermediate frames.
   We employ the REDS\_VTSR dataset derived from diverse videos captured in a hand-held camera for training and evaluation purposes.
   The competition had 62 registered participants, and a total of 8 teams competed in the final testing phase.
   The challenge winning methods achieve the state-of-the-art in video temporal super-resolution.
\end{abstract}

    \let\thefootnote\relax\footnotetext{S. Nah (seungjun.nah@gmail.com, Seoul National University), S. Son, R. Timofte, K. M. Lee are the AIM 2019 challenge organizers, while the other authors participated in the challenge. 
    \\Appendix~\ref{sec_appendix} contains the authors' teams and affiliations.
    \\AIM 2019 webpage:~\url{http://www.vision.ee.ethz.ch/aim19/}}

\section{Introduction}
\label{sec_introduction}

    Video frame interpolation is a classical computer vision task increasing the frame rate of videos. Due to the low-light condition, limited performance of mobile processors, sensors and storage, etc., videos that are recorded in low-frame-rate sometimes exhibit unnatural temporal discontinuity. Video frame interpolation aims to generate the missing views between the recorded frames. For such purpose, video frame interpolation has been adopted in broad fields such as frame rate up-conversion~\cite{castagno1996method, jeon2003coarse, lee2003weighted, choi2007motion, kang2007motion} and video coding~\cite{huang2011multi}. Other applications include video slow-motion effect~\cite{Jiang_2018_CVPR} and novel view synthesis~\cite{stich2008view, Flynn_2016_CVPR}.
    
    As machine learning has been a recent trend in computer vision, deep neural networks are achieving great successes in video frame interpolation~\cite{long2016learning,Liu_2017_ICCV, Niklaus_2017_CVPR,Niklaus_2017_ICCV,Niklaus_2018_CVPR, hu2017multi, xue2019video, Jiang_2018_CVPR, Meyer_2018_CVPR, Peleg_2019_CVPR, Bao_2019_CVPR}.
    However, there has been no standard dataset for video frame interpolation, and most of the existing methods are trained from different datasets. Often, the videos are self-collected from Flickr~\cite{Niklaus_2017_CVPR} or YouTube~\cite{Niklaus_2017_ICCV,Niklaus_2018_CVPR,Jiang_2018_CVPR,Peleg_2019_CVPR}. Other methods propose their own datasets~\cite{xue2019video, MEMC-Net} or use exiting datasets introduced for other purposes such as autonomous driving~\cite{Geiger2013IJRR}, action recognition~\cite{soomro2012ucf101}, deblurring~\cite{Nah_2017_CVPR,Su_2017_CVPR}, and video segmentation~\cite{Perazzi2016,Pont-Tuset_arXiv_2017}.
    
    Thus, in spite of the advances in video frame interpolation research, benchmarking and comparing different methods remain as a nontrivial issue. 
    Furthermore, current widely used benchmark datasets such as Middlebury~\cite{Baker:IJCV:11}, UCF101~\cite{soomro2012ucf101}, THUMOS 2015~\cite{THUMOS15}, Vimeo-90K~\cite{xue2019video} are limited in quantity or strength of motion. 
    While SlowFlow~\cite{Janai2017CVPR}, their high frame rate version of Sintel~\cite{Butler:ECCV:2012} and HD dataset~\cite{MEMC-Net} are in higher quality, they are not much popular for frame interpolation benchmark, yet. 
    
    In this paper, we report the AIM 2019 Challenge on Video Temporal Super-Resolution with a focus on the submitted methods and the benchmark results. 
    We provide a new large-scale dataset, REDS\_VTSR to train and evaluate the video temporal super-resolution methods in a unified environment.
    The proposed REDS\_VTSR dataset consists of high-quality dynamic scenes with large motions at different frame rates. The dataset is derived from the superset of videos that are used to create REDS~\cite{Nah_2019_CVPR_Workshops_REDS} dataset for video deblurring~\cite{Nah_2019_CVPR_Workshops_Deblur} and super-resolution~\cite{Nah_2019_CVPR_Workshops_SR}.
    There are 30 sequences each consisting of 181 frames at 60 fps where the corresponding 30 fps and 15 fps counterparts are subsampled from.
    The challenge goal is to recover 60 fps video from low-frame-rate 15 fps input sequence.%
    In the following sections, we describe the related works and introduce the AIM 2019 Video Temporal Super-Resolution challenge~(VTSR).
    We also present and discuss the challenge results with the proposed methods.
    
\section{Related Works}
    \label{sec_related_works}

    Classical image interpolation techniques mostly include motion modeling via optical flow~\cite{Baker:IJCV:11} or phase shift~\cite{didyk2013joint,wadhwa2013phase}. Modern video frame interpolation methods also employ motion estimation stemming from them, developing more sophisticated motion modeling and warping methods. Meanwhile, many video datasets are adopted for training and evaluating those proposed methods.

    \subsection*{Frame interpolation}
    \label{sec_frame_interpolation}

    Long~\etal~\cite{long2016learning} proposed MIND, an early CNN based approach that learns to estimate the interpolated frame without explicit motion modeling. However, their primary goal was to infer the correspondence between input frames by inverting the interpolation network, and the interpolation accuracy was not evaluated.

    Meyer~\etal~\cite{Meyer_2015_CVPR} proposed a phase-based frame interpolation by using the phase difference between input frames without using optical flow. As no direct pixel correspondence is calculated, the method is robust against the illumination change and blur. Later, Meyer~\etal~\cite{Meyer_2018_CVPR} proposed PhaseNet that predicts amplitude and phase decomposition of the intermediate frame. The CNN is trained in an end-to-end manner via image loss and phase loss. 
    
        \begin{table*}[t]
        \centering
        \begin{tabular}{c c c c c}
            Name & Resolution & fps & \# Sequences / \# Frames & Note \\
            \hline
            \hline
            Flickr~\cite{Niklaus_2017_CVPR} & 150 $\times$ 150 & - & 250,000 / 750,000$^\ast$ & \multirow{4}{*}{
                \makecell{
                    $^\ast$The number of training
                    \\ patches are reported
                    \\ instead of the number of
                    \\ original frames.
                }
            } \\
            YouTube~\cite{Niklaus_2017_ICCV} & 150 $\times$ 150 & - & 250,000 / 750,000$^\ast$ \\
            YouTube~\cite{Niklaus_2018_CVPR} & 300 $\times$ 300 & - & 50,000 / 150,000$^\ast$ \\
            YouTube~\cite{Peleg_2019_CVPR} & 512 $\times$ 512 & - & 40,000 / 120,000$^\ast$ \\
            YouTube + Adobe 240fps~\cite{Su_2017_CVPR} & - & - & 1,132 / 376K & Used at SuperSloMo~\cite{Jiang_2018_CVPR} \\
            UCF101~\cite{soomro2012ucf101} & 320 $\times$ 240 & 25 & 13,320 / - & - \\
            Vimeo-90k~\cite{xue2019video} & 448 $\times$ 256 & - & 73,171 / 219,513 & Vimeo-90k triplet dataset \\
            KITTI raw~\cite{Geiger2013IJRR} (downsampled) & 384 $\times$ 128 & 10 & 56 / 16,951 & - \\
            GoPro~\cite{Nah_2017_CVPR} & 1280 $\times$ 720 & - & 33 / 3,214 & - \\
            DAVIS 2016~\cite{caelles2017one} & 720 $\times$ 480 & - & 50 / 3,455 & \multirow{2}{*}{
                \makecell{
                    Used at PhaseNet~\cite{Meyer_2018_CVPR}
                }
            } \\
            DAVIS 2017~\cite{Pont-Tuset_arXiv_2017} & 720 $\times$ 480 & - & 150 / 10,459 \\
            \hline
            \multicolumn{5}{c}{Evaluation~(test) only} \\
            \hline
            Middlebury~\cite{Baker:IJCV:11} & - & 60 & 8 / 58 & - \\
            THUMOS 2015~\cite{THUMOS15, idrees2017thumos} & - & - & 5,613 / - & - \\
            SlowFlow~\cite{Janai2017CVPR} & 1280 $\times$ 1024 & $>$200 & 46 / - & - \\
            Sintel$^\dagger$~\cite{Butler:ECCV:2012, Janai2017CVPR} & 2048 $\times$ 872 & 1008 & 19 / - & Synthetic data \\
            HD dataset~\cite{MEMC-Net} & 1080p, 720p, 544p & - & 11 / - & Large motion \\
            \hline
            \multicolumn{5}{c}{AIM 2019 VTSR challenge} \\
            \hline
            REDS\_VTSR & 1280$\times$720 & 60 & 300 / 54,300 & Dynamic scenes \\
        \end{tabular}
        \caption{
            Frame interpolation dataset statistics. 
            Sintel$^\dagger$~\cite{Butler:ECCV:2012} sequences are rendered in high-frame-rate~\cite{Janai2017CVPR}.
        }
        \vspace{-0.4cm}
        \label{tab:dataset}
    \end{table*}
    
    There were other approaches that tried to unify motion estimation and frame synthesis. Niklaus~\etal~\cite{Niklaus_2017_CVPR} extracted $41\times41$ spatially adaptive convolutional kernel from a CNN model. The kernel is convolved with the input to synthesize the intermediate frame. Niklaus~\etal~\cite{Niklaus_2017_ICCV} factorized the spatial kernel with 1D kernels to reduce memory footprint. 
    
    On the other hand, the concept of optical flow was often embedded in neural network architectures. 
    Liu~\etal~\cite{Liu_2017_ICCV} implements a voxel flow layer where the spatial component is the optical flow at the corresponding time. By assuming the flow to be locally linear and temporally symmetric, the following volume sampling layer predicts the target frame. 
    Niklaus~\etal~\cite{Niklaus_2018_CVPR} relaxes the assumptions by using bidirectional flow~\cite{Sun_2018_CVPR} to warp the input frames and features. 
    
    However, optical flow estimation could bring errors in case of occlusion. 
    Xue~\etal~\cite{xue2019video} implicitly handles occlusion by learning task-specific flow followed by spatial transformer~\cite{Jaderberg_2015_NIPS} for the area where warping from the optical flow may fail.
    In contrast, Jiang~\etal~\cite{Jiang_2018_CVPR} introduces a soft visibility map for occlusion reasoning to calculate the contribution of the corresponding inputs to the target time. Also, the flow interpolation network approximates the flow from the arbitrary intermediate time frame to inputs. Thus, the proposed SuperSloMo method could directly generate the intermediate frame at any intermediate moment without recursion in contrast to other methods that estimate the middle frame only.
    Hu et al.~\cite{hu2017multi} also proposed an anytime estimation model by unifying frame interpolation and extrapolation framework by applying transitive consistency loss.
    
    Bao et al.~\cite{Bao_2019_CVPR} combines the kernel and optical flow based approaches. Their adaptive warping layer synthesizes a new pixel value by applying a local convolutional kernel where the position of the kernel window is determined by optical flow. During the warping process, relative monocular depth is estimated so that closer objects contribute more during flow projection. 
    Moreover, Peleg et al.~\cite{Peleg_2019_CVPR} focuses on real-time processing and a large receptive field. They extract a low-resolution feature from multi-scale architecture to acquire vertical and horizontal motion vector field. The model is trained with kernel loss~\cite{Niklaus_2017_ICCV}, trilinear interpolation loss~\cite{Liu_2017_ICCV}, etc. As their model operates on lower resolution inputs, the receptive field is large, capable of handling motion size up to $192\times192$.

    The AIM 2019 video temporal super-resolution challenge participants mostly adopt the designs of the previous methods or modify the network architecture to improve performance or reduce computational complexity.
    
    \subsection*{Frame Interpolation Datasets}
    \label{sec_datasets}
    
    \noindent \textbf{Middlebury}~\cite{Baker:IJCV:11} is one of the most popular benchmarks for frame interpolation, which was originally designed for optical flow evaluation. While it is widely used, it has only 8 scenes, limited in quantity. The metric is IE (interpolation error) which can be directly converted to PSNR.
    
    \noindent \textbf{UCF101}~\cite{soomro2012ucf101} is a dataset designed for action classification. It contains various dynamics such as human-object interaction, sports, playing instruments. The training set was used for Deep Voxel Flow~\cite{Liu_2017_ICCV} while the test set is more often employed~\cite{Liu_2017_ICCV,Jiang_2018_CVPR,xue2019video,hu2017multi,Bao_2019_CVPR}. The resolution is $320\times240$.
    
    \noindent \textbf{THUMOS 2015}~\cite{idrees2017thumos} is also a dataset for action recognition whose videos are collected from YouTube. The test set contains 5613 videos of 101 action classes that are compatible with UCF101. It is used for evaluation together with the UCF101 dataset in \cite{Liu_2017_ICCV,hu2017multi,xue2019video}.

    \noindent \textbf{Vimeo-90k}~\cite{xue2019video} consist of $448\times256$frame triplets extracted from video clips collected from Vimeo. From the originally collected pool of videos, static scenes and scenes with large illumination changes are removed. Also, nonlinear motions are also not included. However, this makes the frame interpolation task to be easier than the real-world scenario. It is used for training and testing TOF~\cite{xue2019video}, DAIN~\cite{Bao_2019_CVPR}. IM-Net~\cite{Peleg_2019_CVPR} uses super-resolved Vimeo sequences to evaluate the interpolation performance for large motion.
    
    \noindent \textbf{SlowFlow}~\cite{Janai2017CVPR} are high quality dataset for optical flow. The videos are recorded with a high-speed camera at $2560\times1440$ resolution and $>200$ fps. The scenes are diverse and contain varying levels of realistic blur. The \textbf{MPI Sintel}~\cite{Butler:ECCV:2012} are re-rendered at $1008$ fps and $2048\times872$ resolution to check optical flow methods. They are adopted for evaluating the slow-motion algorithm~\cite{Jiang_2018_CVPR}. 
    
    \noindent \textbf{DAVIS} datasets~\cite{Perazzi2016,Pont-Tuset_arXiv_2017} are benchmark datasets for video object segmentation with a natural level of motion blur, appearance changes, camera shake, etc. They are used to train PhaseNet~\cite{Meyer_2018_CVPR}.
    
    \noindent \textbf{KITTI raw}~\cite{Geiger2013IJRR} is a dataset for autonomous driving captured on a car, recorded in 5 categories: Road, City, Residential, Campus and Person. There are 56 sequences where Long et al.~\cite{long2016learning} downsampled to $384\times128$. They additionaly train on downsampled Sintel~\cite{Butler:ECCV:2012} at $256\times128$.
    
    \noindent \textbf{HD} dataset~\cite{MEMC-Net} are 7 video clips collected from Xiph website each consisting of 50 frames and 4 clips from Sintel~\cite{Butler:ECCV:2012}. The evaluation was done in \cite{MEMC-Net,Bao_2019_CVPR}.
    
    \noindent \textbf{GOPRO} dataset~\cite{Nah_2017_CVPR} is a dataset captured with a high-speed camera at 240 fps for dynamic scene deblurring. The captured frames are averaged to synthesize blurry images with reference sharp frames.  MSFSN~\cite{hu2017multi} is trained with it.
    
    \noindent \textbf{Adobe 240fps}~\cite{Su_2017_CVPR} dataset is also a deblurring dataset recorded with high-speed cameras. SuperSloMo~\cite{Jiang_2018_CVPR} used it jointly with YouTube videos for training. 
    
    \noindent \textbf{Others} Several methods collected videos from the web to train their own methods. Different sets of YouTube~\cite{Niklaus_2017_ICCV,Niklaus_2018_CVPR,Jiang_2018_CVPR,Peleg_2019_CVPR} and Flickr~\cite{Niklaus_2017_CVPR} videos are collected for each method.
    
    
    

    As every method uses different training and test datasets, it is difficult to perform a fair comparison between them. The dataset statistics are summarized in Table~\ref{tab:dataset}.
    Also, many of the datasets lack quality in terms of resolution, diversity of motion, or quantity. In contrast, our proposed REDS\_VTSR dataset contains various dynamic motion of objects and a camera. Further, we evaluate the challenge participants' methods in a unified environment.
    
    \begin{figure*}[t]
        \centering

        \subfloat[Middlebury~\cite{Baker:IJCV:11}]{\includegraphics[width=0.48\textwidth]{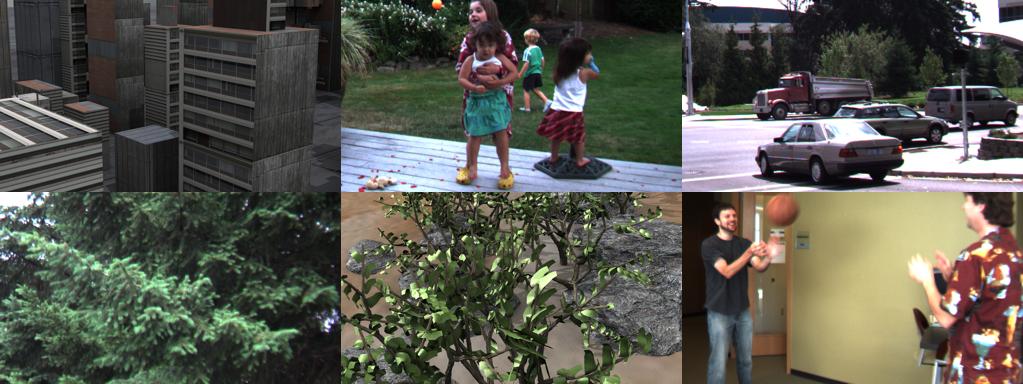}}\hspace{0.02\textwidth}%
		\subfloat[UCF101~\cite{soomro2012ucf101}]{\includegraphics[width=0.48\textwidth]{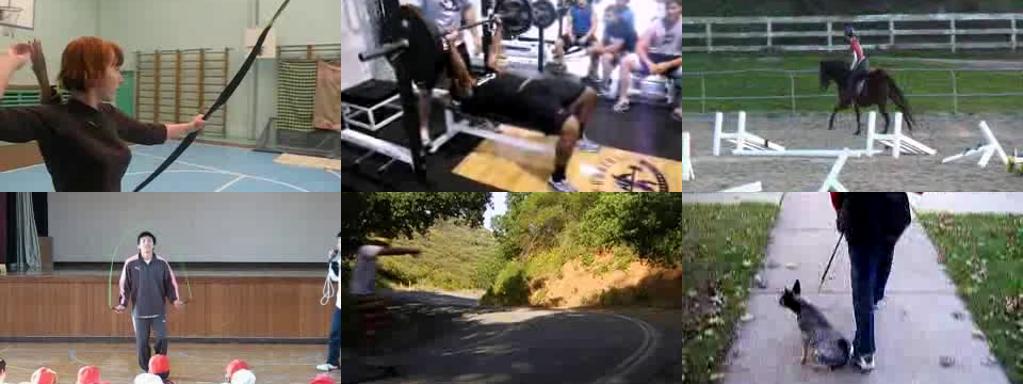}}
		\vspace{-0.3cm}
		\\
		\subfloat[Vimeo-90k~\cite{xue2019video}]{\includegraphics[width=0.48\textwidth]{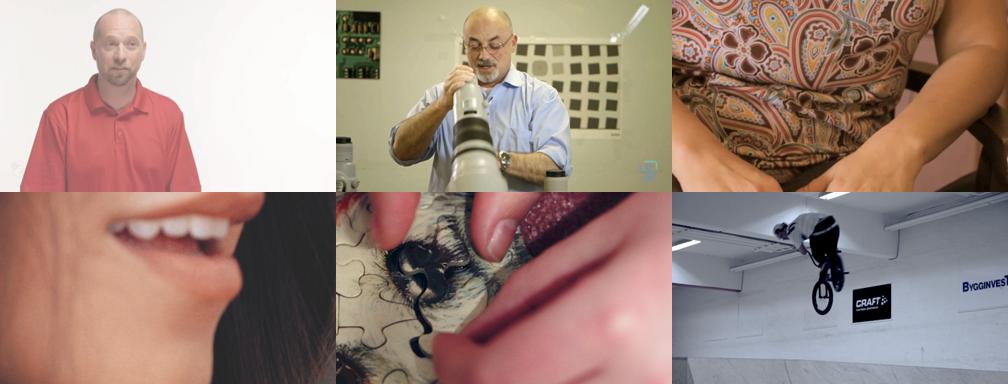}}\hspace{0.02\textwidth}%
		\subfloat[SlowFlow~\cite{Janai2017CVPR}]{\includegraphics[width=0.48\textwidth]{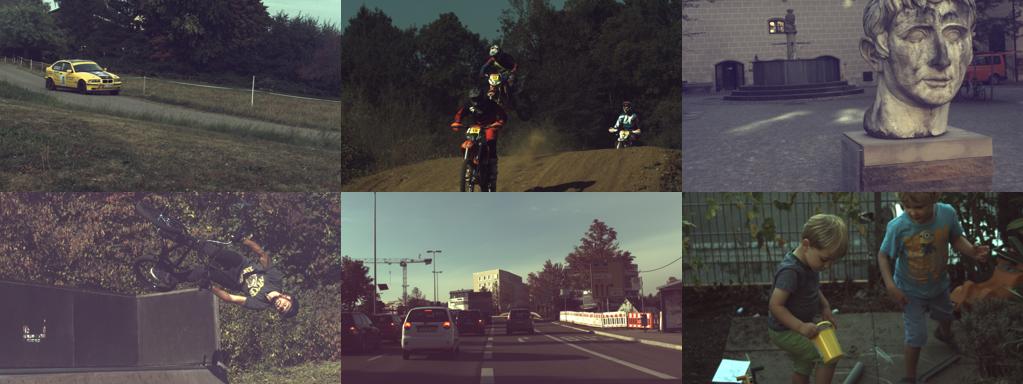}}
		\vspace{-0.3cm}
		\\
		\subfloat[Adobe 240fps~\cite{Su_2017_CVPR}]{\includegraphics[width=0.48\textwidth]{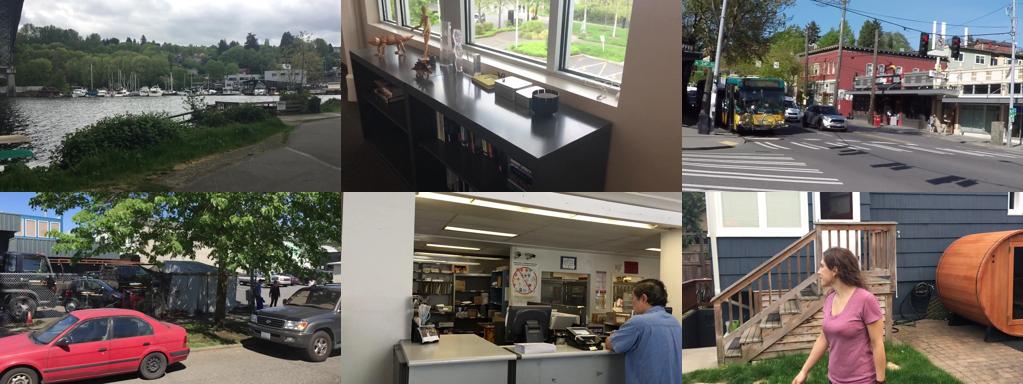}}\hspace{0.02\textwidth}%
		\subfloat[GoPro~\cite{Nah_2017_CVPR}]{\includegraphics[width=0.48\textwidth]{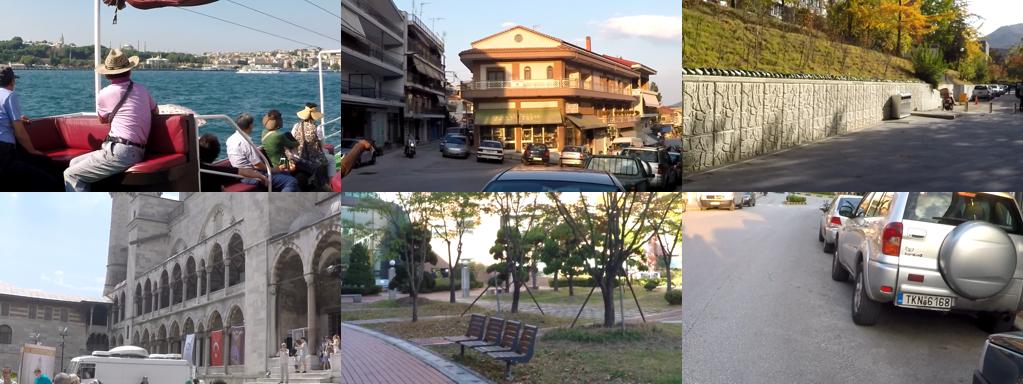}}
		\vspace{-0.3cm}
		\\
		\subfloat[Sintel~\cite{Butler:ECCV:2012}]{\includegraphics[width=0.48\textwidth]{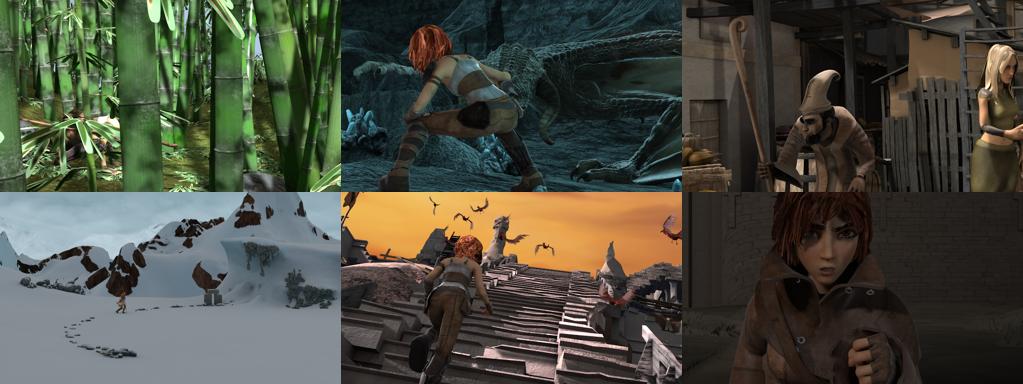}}\hspace{0.02\textwidth}%
		\subfloat[REDS\_VTSR]{\includegraphics[width=0.48\textwidth]{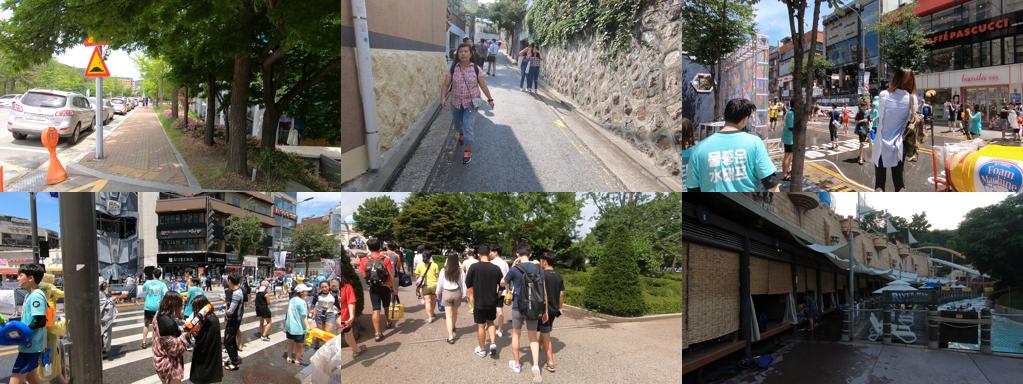}}
        \vspace{-0.3cm}
        \\

        \caption{
            Example frames from selected datasets.
            All samples are cropped to $16:9$ ratio for better visualization.
        }
        \label{fig:dataset}
        \vspace{-0.5cm}
    \end{figure*}
    
    \begin{figure*}[h]
        \centering
        \includegraphics[width=\textwidth]{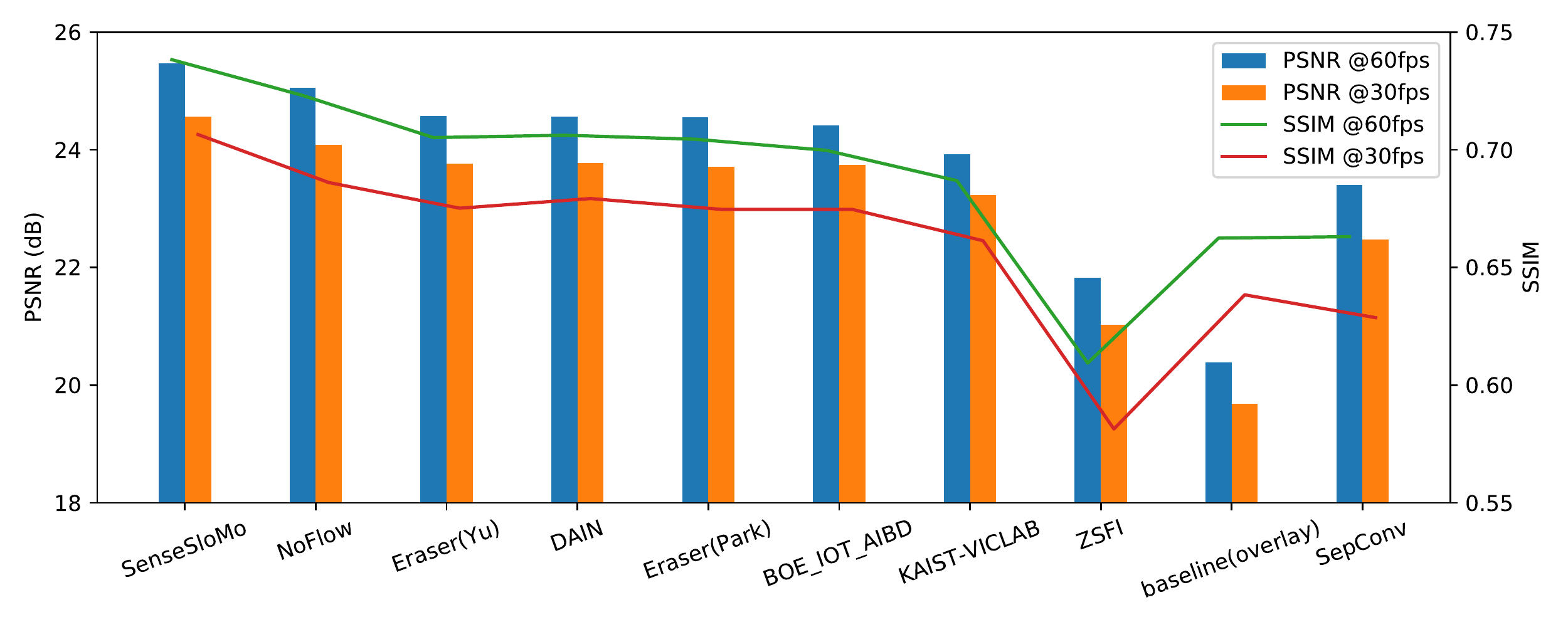}
        \vspace{-0.8cm}
        \caption{Test PSNR and SSIM of the solutions in 60 and 30 fps restoration.}
        \label{fig:result}
        \vspace{-0.1cm}
    \end{figure*}
    
    \begin{table*}[h]
        \begin{center}
            \newcommand{\mc}{\makecell}
            \begin{tabular}{l l| c c| c c}
                & & \multicolumn{2}{c}{15fps $\rightarrow$ 60fps} & \multicolumn{2}{c}{15fps $\rightarrow$ 30fps}\\
                Team & Author & \mc{PSNR} & \mc{SSIM} & \mc{PSNR} & \mc{SSIM} \\
                \hline
                \hline
                SenseSloMo & Siyao & \mc{25.47} & \mc{0.7383}  & \mc{24.56} & \mc{0.7065} \\
                NoFlow & myungsub & \mc{25.05} & \mc{0.7231} & \mc{24.08} & \mc{0.6861} \\
                Eraser & Songsaris & \mc{24.58} & \mc{0.7052} & \mc{23.77} & \mc{0.6752} \\
                DAIN & wangshen233 & \mc{24.56} & \mc{0.7062} & \mc{23.78} & \mc{0.6793} \\
                Eraser & BumjunPark & \mc{24.55} & \mc{0.7045} & \mc{23.71} & \mc{0.6747} \\
                BOE\_IOT\_AIBD\_IMP & BOE\_IOT\_AIBD\_IMP & \mc{24.41} & \mc{0.6998} & \mc{23.74} & \mc{0.6747} \\
                KAIST-VICLAB & WSPark & \mc{23.93} & \mc{0.6869} & \mc{23.23} & \mc{0.6614} \\
                ZSFI & gpkoko & \mc{21.83} & \mc{0.6094} & \mc{21.03} & \mc{0.5814} \\
                \hline
                SepConv~\cite{Niklaus_2017_ICCV} & - & \mc{23.40} & \mc{0.6631} & \mc{22.48} & \mc{0.6287}\\
                \textit{baseline} (overlay) & - & \mc{20.39} & \mc{0.6625} & \mc{19.68} & \mc{0.6384} \\
            \end{tabular}
        \end{center}
        \caption{AIM 2019 Video Temporal Super-Resolution Challenge results on the REDS\_VTSR test data. Teams are sorted by ranking in terms of PSNR for 60 fps restoration.}
        \label{table_results}
    
        \centering
        \begin{tabular}{l| r c c c}
            Team & Runtime~(s) & Platform & \makecell{GPU \\ (at runtime)}  & \makecell{Ensemble / Fusion \\ (at runtime)} \\
            \hline
            \hline
            SenseSloMo & 1.00 & PyTorch & GTX 1060 & flip (x4) \\
            NoFlow & 0.30 & PyTorch & TITAN Xp & - \\
            Eraser~(Yu) & 0.46 & PyTorch & TITAN V & rotation / flip (x8) \\
            DAIN & 0.82 & PyTorch / ATen & RTX 2080 Ti & - \\
            Eraser~(Park) & 2.40 & PyTorch & RTX 2080 Ti & rotation / flip (x8) \\
            BOE\_IOT\_AIBD\_IMP & 0.26 & PyTorch / C++ & Tesla V100 & - \\
            KAIST-VICLAB & 0.25 & PyTorch & TITAN Xp & - \\
            ZSFI & 0.50 & TensorFlow & GTX 1080 Ti / RTX 2080 & - \\
        \end{tabular}
        \caption{
            Reported runtime per frame on REDS\_VTSR test data (60fps) and details from the factsheets.
            We note that the ZSFI team's method requires 1-3 hour sequence-specific training.
        }
        \label{table_details}
    \end{table*}
    
\section{AIM 2019 VTSR Challenge}
    \label{sec_challege}
    
    \noindent \textbf{Challenge Goal}
    The AIM 2019 challenge on video temporal super-resolution is the first challenge of its kind.
    %
    The purpose of the VTSR challenge is to gauge the state-of-the-art in video frame interpolation and facilitate the comparison of different solutions with a single large-scale dataset, REDS\_VTSR.
    %
    We show examples of the previous video datasets and the REDS\_VTSR dataset in Figure~\ref{fig:dataset}.
    
    The challenge objective is to recover higher-frame-rate video sequences from low-frame-rate input sequences. Given the input 15 fps videos, there were two-staged target frame rates of 30 fps and 60 fps. During the online submission period, participants submitted part of the 30 fps recovery results (1/4) due to the limitation of the evaluation server. At the final email submission period, Full 60 fps results were submitted. The results in Table~\ref{table_results} show the restoration performance on the full test dataset at the two frame rates.

    We provide the REDS\_VTSR dataset for the participants to train, validate, and test the performance. The training and validation sets are derived from the same videos as REDS~\cite{Nah_2019_CVPR_Workshops_REDS} dataset while the test set is from a new set of videos. Each of the training, validation, and test set contains 240, 30, 30 sequences, and each sequence has 181 frames at 60 fps. The original videos are captured at 120 fps, and the 15, 30, 60 fps version of them are temporally subsampled from original videos. 
    
    To suppress the artifacts that come from high-speed recording and to avoid redundancies from REDS, we follow a similar process as \cite{Nah_2019_CVPR_Workshops_REDS}. We downsample the $1920\times1080$ frames at $3/4$ ratio and center-crop to make it in standard HD resolution $1280\times720$. As the scene scale is relatively larger than REDS, it facilitates a relatively challenging configuration for frame interpolation with a larger strength of motion. Each frame is saved without compression.

\section{Challenge Results}

    The challenge had 62 registered participants and 8 teams competed in the final test phase. The teams submitted their final 60 fps results, source code, trained models, and factsheets. All the results were reproducible from the submitted source code. We evaluated the solutions in terms of PSNR and SSIM. The competition results are summarized in Table~\ref{table_results} and the implementation specifications are in Table~\ref{table_details}. We visualize the relative performance in Figure~\ref{fig:result}.
    
\begin{figure*}[t]
        \centering
        \newcommand{\w}{0.188\textwidth}
        \newcommand{\wm}{0.010\textwidth}
        \includegraphics[width=0.98\textwidth]{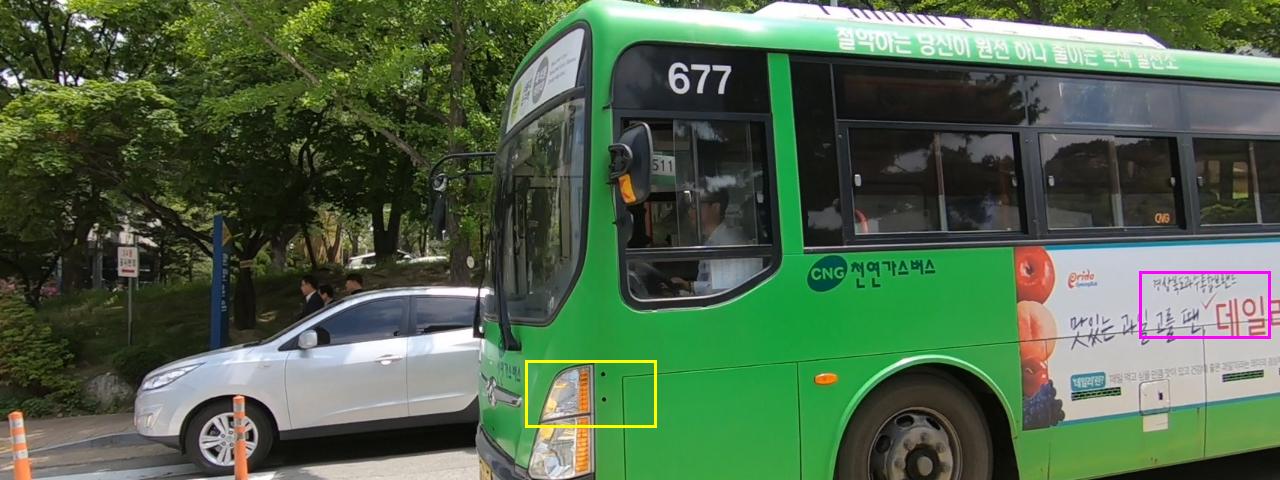}
        \\
        \vspace{0.1cm}
        \includegraphics[width=\w]{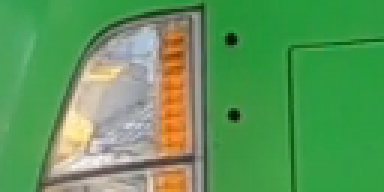}\hspace{\wm}%
        \includegraphics[width=\w]{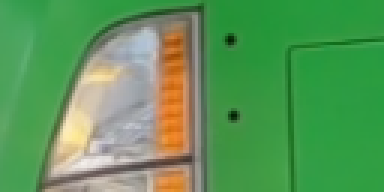}\hspace{\wm}%
        \includegraphics[width=\w]{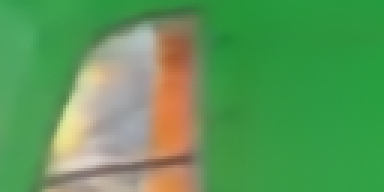}\hspace{\wm}%
        \includegraphics[width=\w]{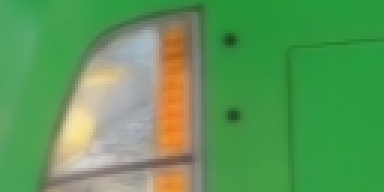}\hspace{\wm}%
        \includegraphics[width=\w]{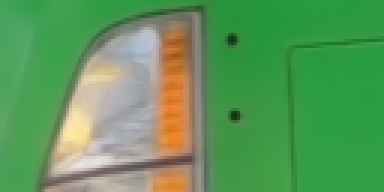}
        \\
        \vspace{-0.25cm}
        \subfloat[GT]{\includegraphics[width=\w]{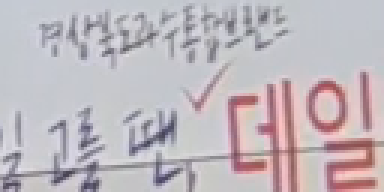}}\hspace{\wm}%
        \subfloat[SenseSloMo]{\includegraphics[width=\w]{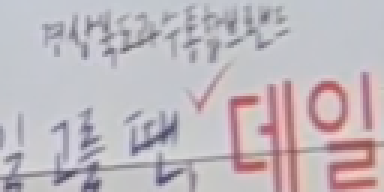}}\hspace{\wm}%
        \subfloat[NoFlow]{\includegraphics[width=\w]{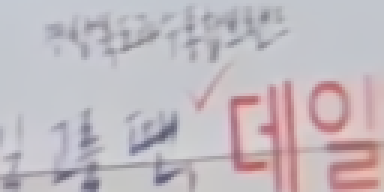}}\hspace{\wm}%
        \subfloat[Eraser~(Yu)]{\includegraphics[width=\w]{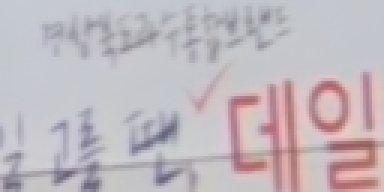}}\hspace{\wm}%
        \subfloat[DAIN]{\includegraphics[width=\w]{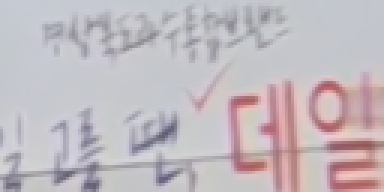}}
        \vspace{-0.1cm}
        \caption{Temporal super-resolution results~(REDS\_VTSR `002/00000006')}
        \label{fig:example_01}
        \vspace{-0.5cm}
    \end{figure*}

    Most of the participating teams used optical flow in their deep CNN architecture. The challenge winner, SenseSloMo~\cite{qvi_iccvw19} modeled nonlinear motion between frames with the quadratic model. They additionally use perceptual loss~\cite{johnson2016perceptual} to improve visual quality. On the other hand, NoFlow avoided using optical flow and chose to use channel attention. 
    Interestingly, the ZSFI team uses a zero-shot approach that does not require training samples. The results of high-ranking teams are compared in Figure~\ref{fig:example_01}.
    

\section{Challenge Methods and Teams}

We describe the submitted solution details in this section.
PyTorch has served as a generally favorable framework while some methods have leveraged customized C++ or ATen kernels for flow prediction.
Several methods are reporting that run-time self-ensemble strategy~\cite{Timofte_2016_CVPR} can enhance the model performance at the cost of running speed.

\begin{figure}[h]
    \begin{center}
        \includegraphics[width=\linewidth]{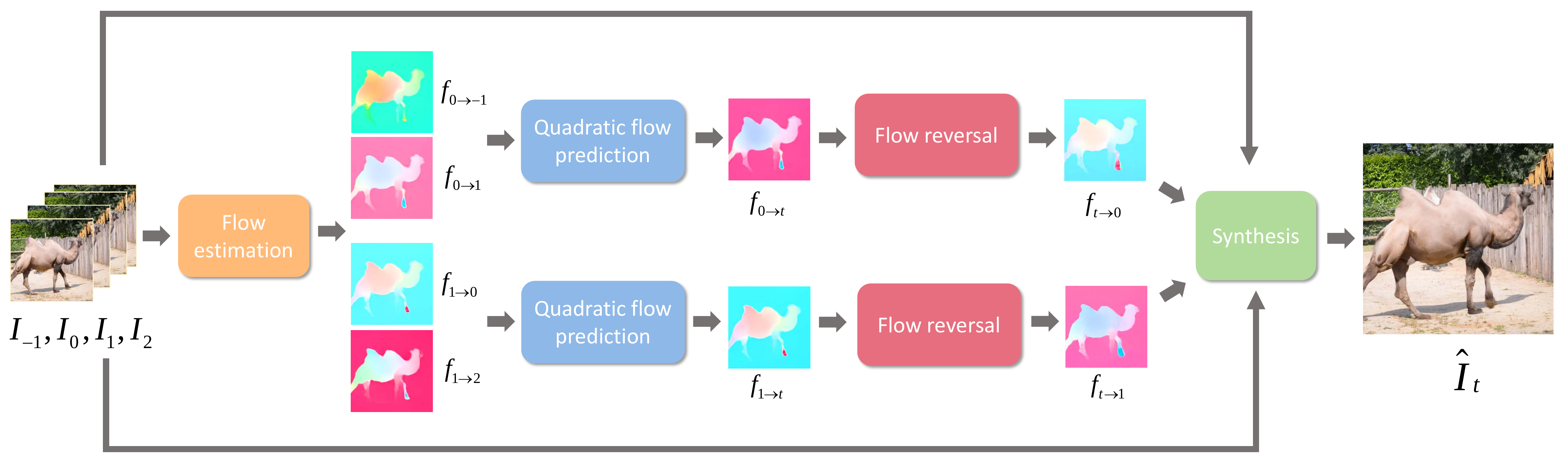}
        \caption{
            SenseSloMo: the proposed quadratic frame interpolation method.
        }
        \label{fig:senseslomo}
        \vspace{-0.65cm}
    \end{center}
\end{figure}

\subsection{SenseSloMo}

The SenseSloMo team takes a quadratic frame interpolation method~\cite{qvi_nips19,qvi_iccvw19} which can reflect the acceleration of motions among video frames as in Figure~\ref{fig:senseslomo}.
Specifically, the proposed method uses four input frames $I_{-1}, I_0, I_1, I_2$ to generate an arbitrary inter frame $I_t \left( 0 < t < 1 \right)$.
Generally, the proposed model can be divided into four parts.
First, the network estimates optical flows $F_{0 \rightarrow {-1}}, F_{0 \rightarrow 1}, F_{1 \rightarrow 0}$ and $F_{1 \rightarrow 2}$.
Second, $F_{0 \rightarrow t}$ and $F_{1 \rightarrow t}$ are calculated using a quadratic flow formula.
Third, the SenseSloMo team reverses those flows to $F_{t \rightarrow 0}$ and $F_{t \rightarrow 1}$.
Finally, the interpolated frame $I_t$ is synthesized.

\subsection{NoFlow}

\begin{figure}[b]
    \begin{center}
        \includegraphics[width=\linewidth]{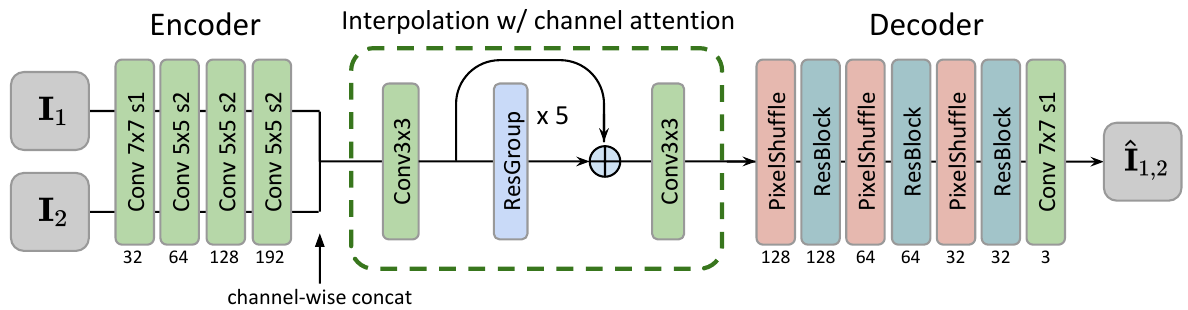}
        \caption{NoFlow: overall network architecture.}
        \label{fig:noflow_1}
        \vspace{-0.5cm}
    \end{center}
\end{figure}

The NoFlow team is motivated by the potential drawbacks in using optical flow and proposes a novel framework of video frame interpolation that replaces optical flow with a simple convolutional network as in Figure~\ref{fig:noflow_1}.
Encoding images to lower spatial resolution gradually distributes the motion-related information into multiple channels to construct a transformed feature map.
This feature representation is then combined with channel attention to capture the variations between the two input frames and synthesizes high-quality intermediate video frames without explicit motion estimation.

\begin{figure}[b]
    \begin{center}
        \includegraphics[width=\linewidth]{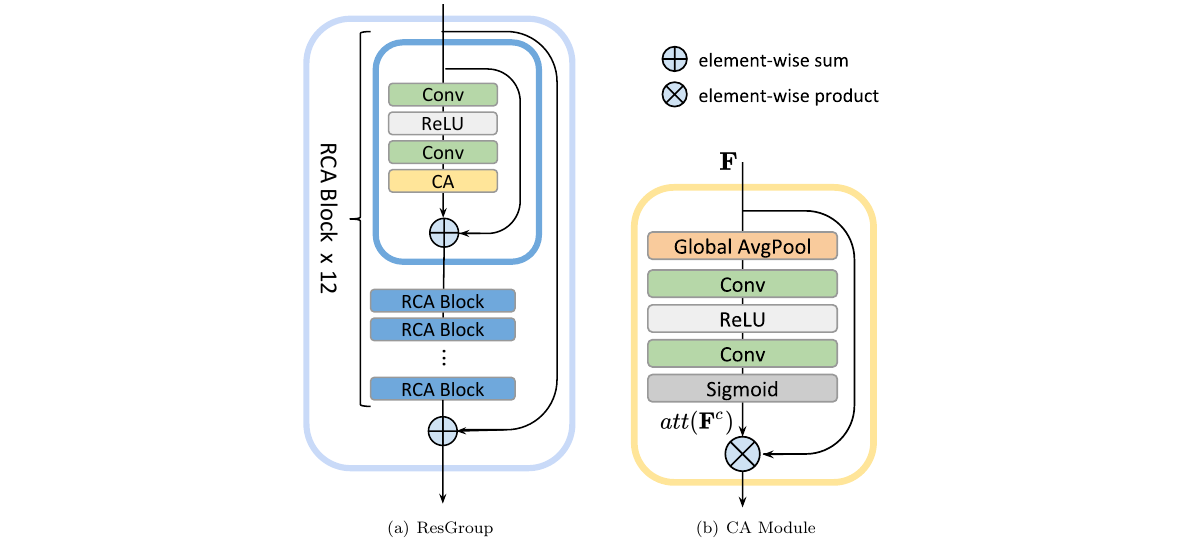}
        \caption{ResGroup and CA modules in NoFlow architecture.}
        \label{fig:noflow_2}
        \vspace{-0.5cm}
    \end{center}
\end{figure}

Using ResGroups and CA modules as in Figure~\ref{fig:noflow_2}, the proposed approach is capable of handling large motion and heavy occlusion effectively and outperforms the prior state-of-the-art methods.

\begin{figure}[b]
    \begin{center}
        \includegraphics[width=\linewidth]{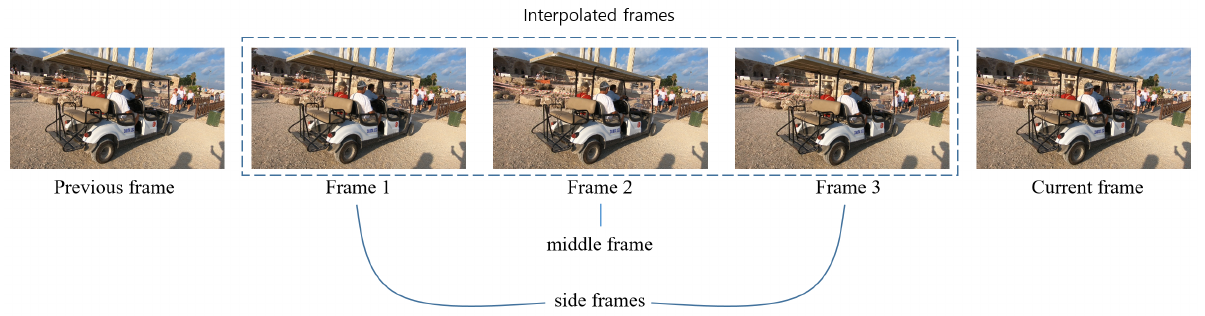}
        \caption{Eraser~(Yu): an example of frame index representation.}
        \label{fig:Eraser_yu_1}
        \vspace{-0.5cm}
    \end{center}
\end{figure}

\begin{figure}[b]
    \begin{center}
        \includegraphics[width=\linewidth]{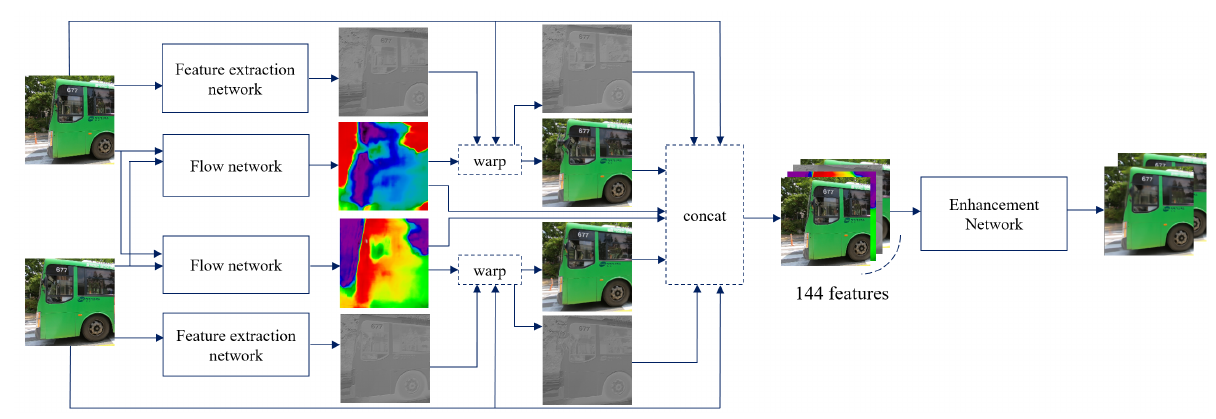}
        \caption{Eraser~(Yu): overall pipeline of the method.}
        \label{fig:Eraser_yu_2}
        \vspace{-0.5cm}
    \end{center}
\end{figure}

\begin{figure}[b]
    \begin{center}
        \includegraphics[width=\linewidth]{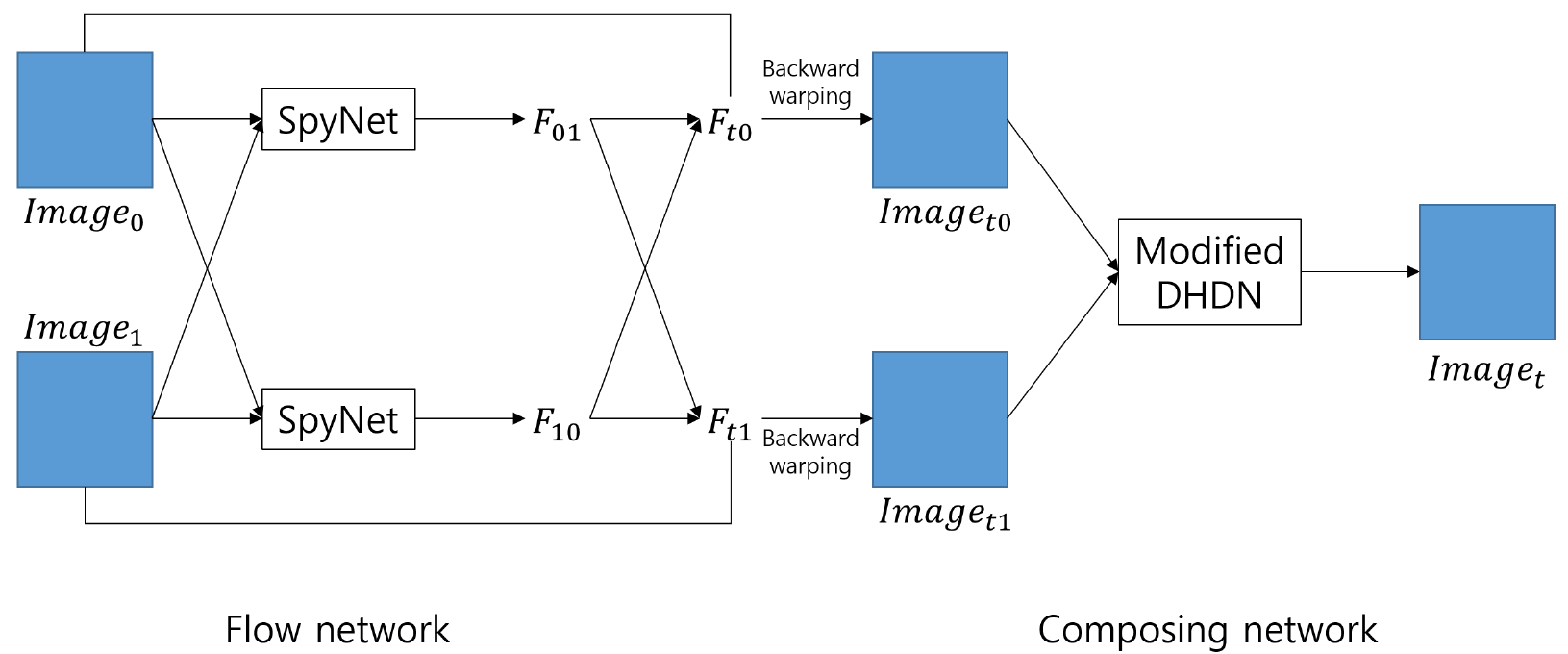}
        \caption{Eraser~(Park): the proposed DVTSR architecture.}
        \label{fig:Eraser_park}
        \vspace{-0.5cm}
    \end{center}
\end{figure}

\subsection{Eraser}

The Eraser team has submitted two solutions(Yu, Park) for video frame interpolation.

Eraser~(Yu) predicts 3 intermediate frames for each frame pair.
The frame index representation used in this approach is shown in Figure~\ref{fig:Eraser_yu_1}.
The proposed method uses two separate models: a middle network to estimate a middle frame~(Frame 2) and a side network to estimate side frames~(Frame 1 and 3).
Each model consists of optical flow estimation and enhancement as Figure~\ref{fig:Eraser_yu_2}.
For initial optical flow estimation, a pre-trained PWC-Net~\cite{Sun_2018_CVPR} is used.
Then a modified DIDN~\cite{Yu_2019_CVPR_Workshops} is introduced as an enhancement network.
In the proposed approach, at least one flow estimator exists for each frame position.
For example, one estimator is used for position 1, two estimators for position
2, and one estimator for position 3.
Because position 2 is in the middle, two estimators (forward direction and backward direction) can be allocated.
Frames at positions 1 and 3 are predicted using only close frames (the previous frame is used for predicting position 1 and current frame for position 3).
As an input of the enhancement network, all of the extracted features from ResNet-151~\cite{He_2016_CVPR}, warped frames from PWCNet, original frames, and flow information is used.
The pre-trained flow estimator is fine-tuned for a specific frame position and is named as a network using position-specific flow~(PoSNet).
All networks (PWC-Net + modified DIDN) are trained together in an end-to-end manner.
In other words, as in ToFlow~\cite{xue2019video}, the pre-trained PWC-Net is also fine-tuned for the challenge dataset.

Eraser~(Park) 
is inspired by TOFlow~\cite{xue2019video} and SuperSloMo~\cite{Jiang_2018_CVPR}.
Figure~\ref{fig:Eraser_park} shows the architecture of the DVTSR.
The model consists of two networks: flow network and composing network.
For the flow network, the Eraser team~(Park) adopts the architecture of SpyNet~\cite{Ranjan_2017_CVPR}.
For the composing network, the architecture of DHDN~\cite{Park_2019_CVPR_Workshops} is used with modification.
The number of initial feature maps is 30, and one DCR block is used per each step.

\subsection{DAIN}

\begin{figure}[h]
    \begin{center}
        \includegraphics[width=\linewidth]{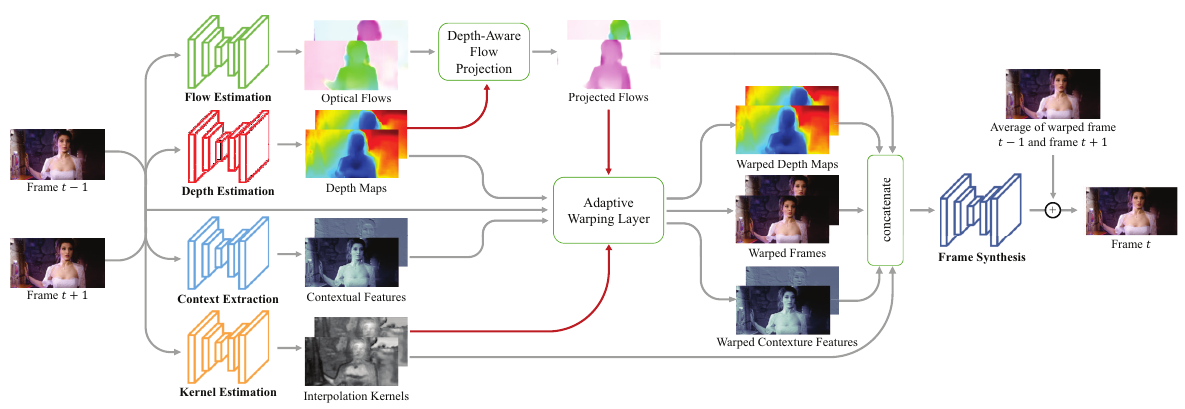}
        \caption{DAIN: overall pipeline of the method.}
        \label{fig:DAIN}
        \vspace{-0.5cm}
    \end{center}
\end{figure}

The DAIN team proposes a video frame interpolation method, which explicitly detects the occlusion by exploring the depth information.
First, the optical flows and depth maps are extracted from video frames as features.
Then, a depth-aware flow projection layer is used to synthesize intermediate flows that preferably sample closer objects than farther ones.
An adaptive warping layer takes the projected flows along with depth map, encoded contextual features, and interpolation kernel features to produce warped depth maps, warped frames, and warped contextual features together.
At last, the warped features are concatenated and a frame synthesis net output the resulting frame.
The overall pipeline is visualized in Figure~\ref{fig:DAIN}.

\subsection{BOE\_IOT\_AIBD\_IMP}

\begin{figure}[h]
    \begin{center}
        \includegraphics[width=\linewidth]{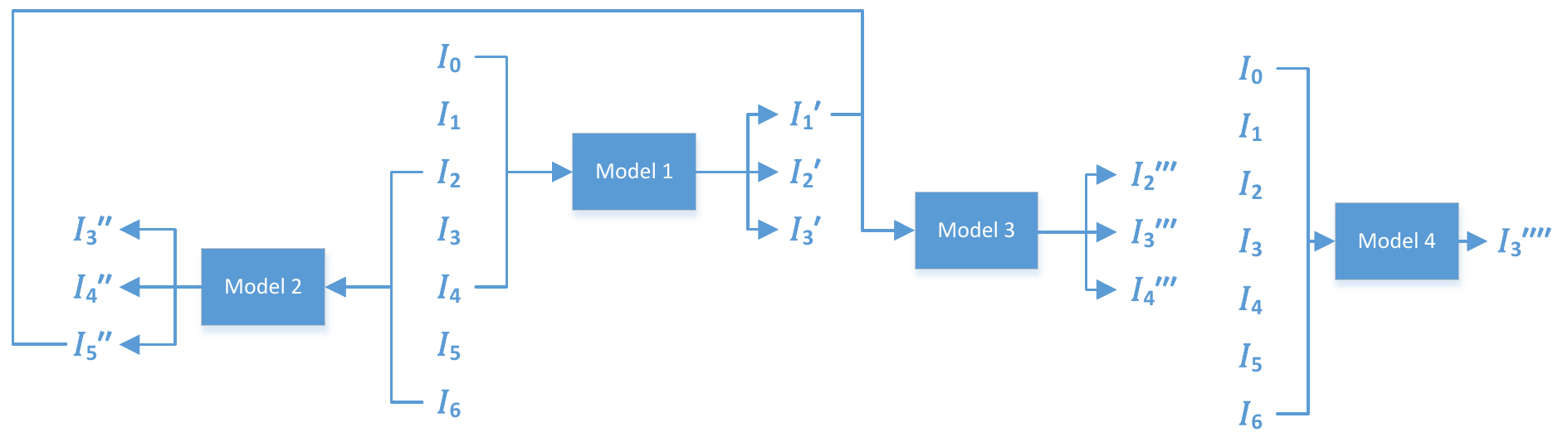}
        \caption{BOE\_IOT\_AIBD\_IMP: the proposed cyclic training process.}
        \label{fig:BOE_1}
        \vspace{-0.5cm}
    \end{center}
\end{figure}

\begin{figure}[h]
    \begin{center}
        \includegraphics[width=\linewidth]{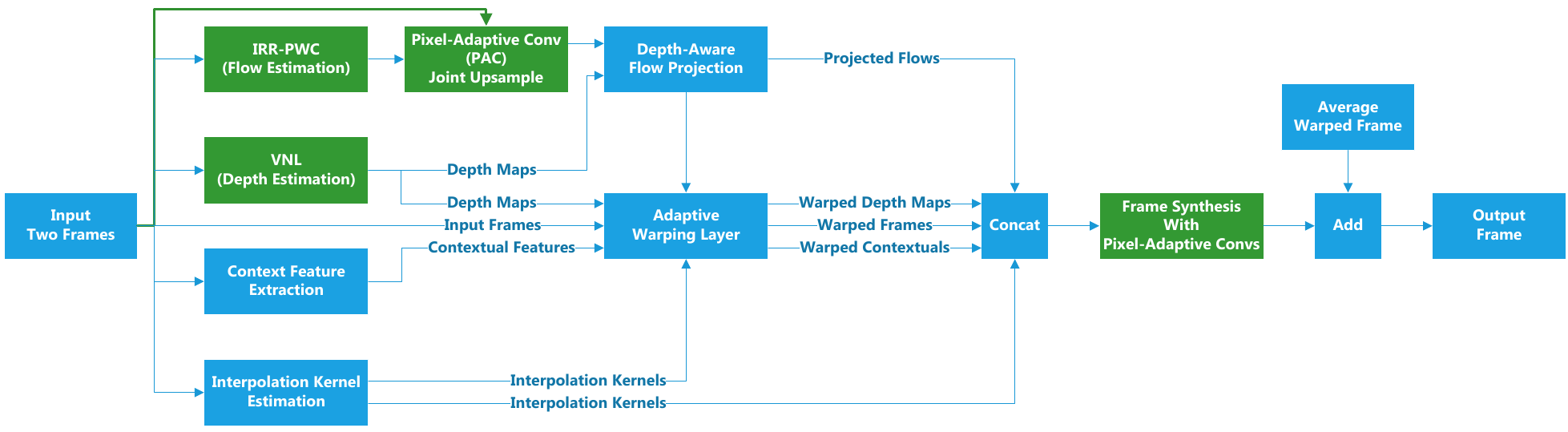}
        \caption{Overall network architecture of the BOE\_IOT\_AIBD\_IMP team.}
        \label{fig:BOE_2}
        \vspace{-0.5cm}
    \end{center}
\end{figure}

The BOE\_IOT\_AIBD\_IMP team has proposed the frame interpolation solution based on DAIN~\cite{Bao_2019_CVPR}.
The team has improved the model by using IRR-PWC~\cite{Hur_2019_CVPR} for calculating the optical flows, PacJointUpsample~\cite{Su_2019_CVPR} for joint optical flow upsampling, VNL~\cite{Yin2019enforcing} for predicting the depth maps and pixel-adaptive convolutions~\cite{Su_2019_CVPR} in frame synthesis module for generating the final results.
Based on CyclicGen~\cite{liu2019cyclicgen}, a new two-stage cyclic generation process is proposed for training the 4x interpolation model as in Figure~\ref{fig:BOE_1}.
The method is using pre-trained IRR-PWC~\cite{Hur_2019_CVPR}, VNL~\cite{Yin2019enforcing}, PacJointUpsample~\cite{Su_2019_CVPR}, and some sub-module/layers of DAIN~\cite{Bao_2019_CVPR}, and was trained/finetuned on the REDS VTSR dataset.
The entire network architecture is visualized in Figure~\ref{fig:BOE_2}.

\subsection{KAIST-VICLAB}

\begin{figure}[h]
    \begin{center}
        \includegraphics[width=\linewidth]{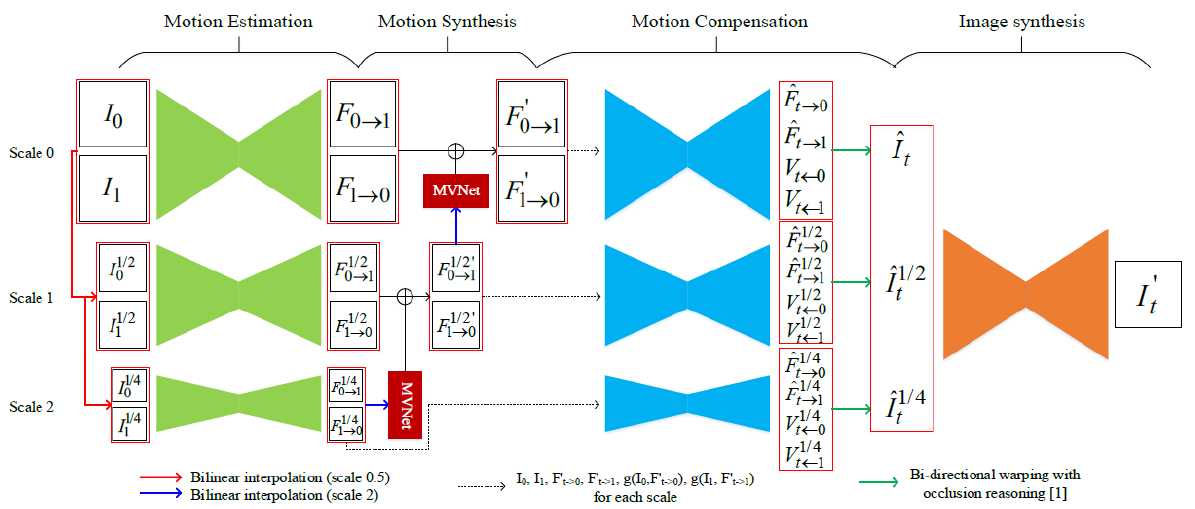}
        \caption{KAIST-VICLAB: overall pipeline of the method.}
        \label{fig:KAIST}
        \vspace{-0.5cm}
    \end{center}
\end{figure}

The KAIST-VICLAB team proposes a method using SuperSloMo~\cite{Jiang_2018_CVPR}, which consists of two U-Nets as the basic structure.
The original SuperSloMo structure estimates the motion vector from a single scale and performs frame interpolation.
The proposed method is mainly composed of four sub-networks, as shown in Figure~\ref{fig:KAIST}.
The first sub-net~(Motion Estimation) estimates the motion vectors
from multi-scale (Scale 0, Scale 1, Scale 2) inputs.
Then, the second sub-net~(Motion Synthesis) synthesizes the motion vectors
in multi-scale (Scale 0, Scale 1, Scale 2) by MVNet.
The third sub-net~(Motion Compensation) performs the frame interpolation for each scale using the estimated multi-scale motion vectors.
Finally, the intermediate frame is obtained through the last sub-network~(Image
synthesis) which finally synthesizes the obtained frames for each scale.
These four sub-networks of our proposed multi-scale structure is trained in an end-to-end manner.
At the second sub-network~(Motion Synthesis), only the residuals of motion vectors for three scales (0, 1, and 2) are calculated by our networks.
Before the last sub-network (Image Synthesis), the reconstructed image for each scale is resized to the original image size by bilinear interpolation.
In Scale 2, our proposed network can process the feature maps with large receptive fields, and in Scale 0, the proposed network can process textured areas efficiently.

\subsection{ZSFI}

\begin{figure}[h]
    \begin{center}
        \includegraphics[width=\linewidth]{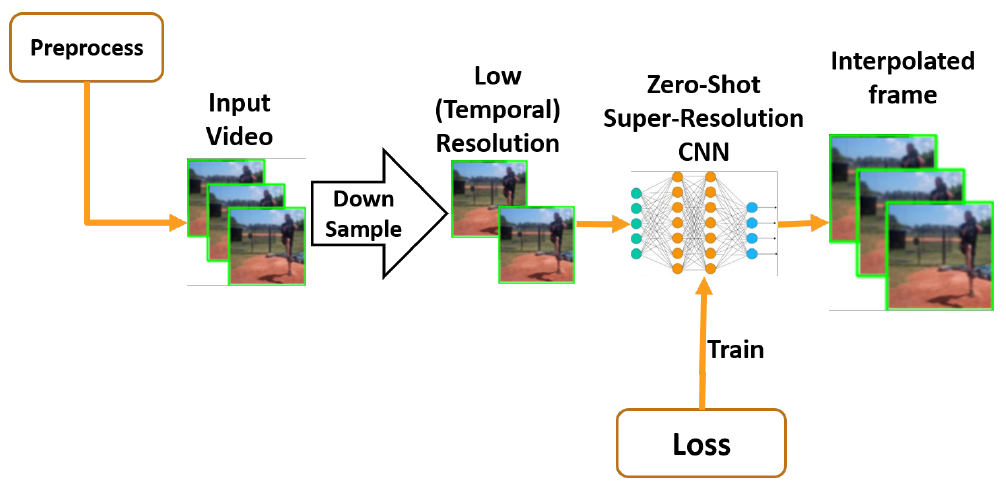}
        \caption{ZSFI: zero-shot training process.}
        \label{fig:ZSFI}
        \vspace{-0.5cm}
    \end{center}
\end{figure}

The ZSFI team proposes a novel ’zero-shot’ frame interpolation, \ie, it reconstructs
and interpolates a video without any prior training.
This approach is based on the concepts of internal statistics and self-similarity in the space-time volume of a video described by Shahar~\etal~\cite{space_time_super}.
The work has demonstrated that space-time patches recur over large distances both in space and in time and most importantly, across scales, that are in different areas of the frame and between frames.
Using the concept of recurring patches, a zero-shot super-resolution algorithm~\cite{Shocher_2018_CVPR} has proposed recently.
Here, the ZSFI team extends this concept to video interpolation from a single example, the video itself as in Figure~\ref{fig:ZSFI}.
From the test video, the ZSFI team creates a temporally down-sampled version of the video that allows us to leverage the internal similarity across scales and learn the interpolation-based solely on the test video.
The proposed approach is especially effective when the target video is significantly different from the training data such as medical or synthetic data.

\section*{Acknowledgments}
We thank the AIM 2019 sponsors.

\appendix
\section{Teams and affiliations}
\label{sec_appendix}
\subsection*{AIM 2019 team}
\noindent\textit{\textbf{Title: }} AIM 2019 Challenge on Video Temporal Super-Resolution \\
\noindent\textit{\textbf{Members: }}
    \textit{Seungjun Nah$^1$ (seungjun.nah@gmail.com)},
    Sanghyun Son$^1$,
    Radu Timofte$^2$,
    Kyoung Mu Lee$^1$\\
\noindent\textit{\textbf{Affiliations: }} \\
$^1$ Department of ECE, ASRI, Seoul National University~(SNU), Korea \\
$^2$ Computer Vision Lab, ETH Zurich, Switzerland \\

\subsection*{SenseSloMo}
\noindent\textit{\textbf{Title: }} Quadratic Video Interpolation \\
\noindent\textit{\textbf{Members: }}
    \textit{Li Siyao (lisiyao1@sensetime.com)},
    Ze Pan,
    Xiangyu Xu,
    Wenxiu Sun \\
\noindent\textit{\textbf{Affiliations: }} \\
SenseTime Research, China \\

\subsection*{NoFlow}
\noindent\textit{\textbf{Title: }} Channel Attention is All You Need for Video Temporal Super-Resolution \\
\noindent\textit{\textbf{Members: }}
    \textit{Myungsub Choi$^1$ (cms6539@gmail.com)},
    Heewon Kim$^1$,
    Bohyung Han$^1$,
    Ning Xu$^2$,
    Kyoung Mu Lee$^1$ \\
\noindent\textit{\textbf{Affiliations: }} \\
$^1$ Department of ECE, ASRI, Seoul National University~(SNU), Korea \\
$^2$ Amazon, USA

\subsection*{Eraser}
\noindent\textit{\textbf{Title: }} Video Frame Interpolation Using Position-Specific Flow \\
\noindent\textit{\textbf{Members: }}
    \textit{Bumjun Park (kkbbbj@gmail.com)},
    Songhyun Yu,
    Sangmin Kim,
    Jechang Jeong \\
\noindent\textit{\textbf{Affiliations: }} \\
Department of ECE, Hanyang University, Korea \\

\subsection*{DAIN}
\noindent\textit{\textbf{Title: }} Depth-Aware Video Frame Interpolation \\
\noindent\textit{\textbf{Members: }}
    \textit{Wang Shen (shenwang@sjtu.edu.cn)},
    Wenbo Bao,
    Guangtao Zhai,
    Li Chen,
    Zhiyong Gao \\
\noindent\textit{\textbf{Affiliations: }} \\
Department of EE, Shanghai Jiao Tong University, China \\

\subsection*{BOE\_IOT\_AIBD\_IMP}
\noindent\textit{\textbf{Title: }} Pixel-Adaptive Joint Depth-aware Cyclic Network for Video Temporal Super-resolution \\
\noindent\textit{\textbf{Members: }}
    \textit{Guannan Chen (578489493@qq.com)},
    Yunhua Lu,
    Ran Duan,
    Tong Liu,
    Lijie Zhang \\
\noindent\textit{\textbf{Affiliations: }} \\
BOE Technology Group Co., Ltd., China \\

\subsection*{KAIST-VICLAB}
\noindent\textit{\textbf{Title: }} Multi-scale Motion Estimation and Motion Compensation \\
\noindent\textit{\textbf{Members: }}
    \textit{Woonsung Park (pys5309@kaist.ac.kr)},
    Munchurl Kim \\
\noindent\textit{\textbf{Affiliations: }} \\
Korea Advanced Institute of Science and Technology~(KAIST), Korea \\

\subsection*{ZSFI}
\noindent\textit{\textbf{Title: }} Zero Shot Frame Interpolation \\
\noindent\textit{\textbf{Members: }}
    \textit{George Pisha (pisha@campus.technion.ac.il)},
    Eyal Naor,
    Lior Aloni \\
\noindent\textit{\textbf{Affiliations: }} \\
Technion Israel Institute of Technology, Israel \\

{\small
\bibliographystyle{ieee}

\begin{thebibliography}{10}\itemsep=-1pt

\bibitem{Baker:IJCV:11}
S.~Baker, D.~Scharstein, J.~P. Lewis, S.~Roth, M.~J. Black, and R.~Szeliski.
\newblock A database and evaluation methodology for optical flow.
\newblock {\em International Journal of Computer Vision (IJCV)}, 92(1):1--31,
  Mar. 2011.

\bibitem{Bao_2019_CVPR}
W.~Bao, W.-S. Lai, C.~Ma, X.~Zhang, Z.~Gao, and M.-H. Yang.
\newblock Depth-aware video frame interpolation.
\newblock In {\em The IEEE Conference on Computer Vision and Pattern
  Recognition (CVPR)}, Jun. 2019.

\bibitem{MEMC-Net}
W.~Bao, W.-S. Lai, X.~Zhang, Z.~Gao, and M.-H. Yang.
\newblock {MEMC}-{Net}: Motion estimation and motion compensation driven neural
  network for video interpolation and enhancement.
\newblock {\em arXiv preprint arXiv:1810.08768}, 2018.

\bibitem{Butler:ECCV:2012}
D.~J. Butler, J.~Wulff, G.~B. Stanley, and M.~J. Black.
\newblock A naturalistic open source movie for optical flow evaluation.
\newblock In {\em The European Conference on Computer Vision (ECCV)}, Oct.
  2012.

\bibitem{caelles2017one}
S.~Caelles, K.-K. Maninis, J.~Pont-Tuset, L.~Leal-Taix{\'e}, D.~Cremers, and
  L.~Van~Gool.
\newblock One-shot video object segmentation.
\newblock In {\em The IEEE conference on Computer Vision and Pattern
  Recognition (CVPR)}, 2017.

\bibitem{castagno1996method}
R.~Castagno, P.~Haavisto, and G.~Ramponi.
\newblock A method for motion adaptive frame rate up-conversion.
\newblock {\em IEEE Transactions on circuits and Systems for Video Technology},
  6(5):436--446, 1996.

\bibitem{choi2007motion}
B.-D. Choi, J.-W. Han, C.-S. Kim, and S.-J. Ko.
\newblock Motion-compensated frame interpolation using bilateral motion
  estimation and adaptive overlapped block motion compensation.
\newblock {\em IEEE Transactions on Circuits and Systems for Video Technology},
  17(4):407--416, 2007.

\bibitem{didyk2013joint}
P.~Didyk, P.~Sitthi-Amorn, W.~Freeman, F.~Durand, and W.~Matusik.
\newblock Joint view expansion and filtering for automultiscopic 3d displays.
\newblock {\em ACM Transactions on Graphics (TOG)}, 32(6):221, 2013.

\bibitem{Flynn_2016_CVPR}
J.~Flynn, I.~Neulander, J.~Philbin, and N.~Snavely.
\newblock Deepstereo: Learning to predict new views from the world's imagery.
\newblock In {\em The IEEE Conference on Computer Vision and Pattern
  Recognition (CVPR)}, Jun. 2016.

\bibitem{Geiger2013IJRR}
A.~Geiger, P.~Lenz, C.~Stiller, and R.~Urtasun.
\newblock Vision meets robotics: The {KITTI} dataset.
\newblock {\em International Journal of Robotics Research (IJRR)}, 2013.

\bibitem{THUMOS15}
A.~Gorban, H.~Idrees, Y.-G. Jiang, A.~Roshan~Zamir, I.~Laptev, M.~Shah, and
  R.~Sukthankar.
\newblock {THUMOS} challenge: Action recognition with a large number of
  classes.
\newblock \url{http://www.thumos.info/}, 2015.

\bibitem{He_2016_CVPR}
K.~He, X.~Zhang, S.~Ren, and J.~Sun.
\newblock Deep residual learning for image recognition.
\newblock In {\em The IEEE Conference on Computer Vision and Pattern
  Recognition (CVPR)}, Jun. 2016.

\bibitem{hu2017multi}
Z.~Hu, Y.~Ma, and L.~Ma.
\newblock Multi-scale video frame-synthesis network with transitive consistency
  loss.
\newblock {\em arXiv preprint arXiv:1712.02874}, 2017.

\bibitem{huang2011multi}
X.~Huang, L.~L. Rak{\^e}t, H.~Van~Luong, M.~Nielsen, F.~Lauze, and
  S.~Forchhammer.
\newblock Multi-hypothesis transform domain wyner-ziv video coding including
  optical flow.
\newblock In {\em 2011 IEEE 13th International Workshop on Multimedia Signal
  Processing}, 2011.

\bibitem{Hur_2019_CVPR}
J.~Hur and S.~Roth.
\newblock Iterative residual refinement for joint optical flow and occlusion
  estimation.
\newblock In {\em The IEEE Conference on Computer Vision and Pattern
  Recognition (CVPR)}, Jun. 2019.

\bibitem{idrees2017thumos}
H.~Idrees, A.~R. Zamir, Y.~Jiang, A.~Gorban, I.~Laptev, R.~Sukthankar, and
  M.~Shah.
\newblock The {THUMOS} challenge on action recognition for videos “in the
  wild”.
\newblock {\em Computer Vision and Image Understanding (CVIU)}, 155:1--23,
  2017.

\bibitem{Jaderberg_2015_NIPS}
M.~Jaderberg, K.~Simonyan, A.~Zisserman, and k.~kavukcuoglu.
\newblock Spatial transformer networks.
\newblock In {\em Advances in Neural Information Processing Systems (NIPS)}.
  2015.

\bibitem{Janai2017CVPR}
J.~Janai, F.~Güney, J.~Wulff, M.~Black, and A.~Geiger.
\newblock {S}low {F}low: Exploiting high-speed cameras for accurate and diverse
  optical flow reference data.
\newblock In {\em The IEEE Conference on Computer Vision and Pattern
  Recognition (CVPR)}, Jun. 2017.

\bibitem{jeon2003coarse}
B.-W. Jeon, G.-I. Lee, S.-H. Lee, and R.-H. Park.
\newblock Coarse-to-fine frame interpolation for frame rate up-conversion using
  pyramid structure.
\newblock {\em IEEE Transactions on Consumer Electronics}, 49(3):499--508,
  2003.

\bibitem{Jiang_2018_CVPR}
H.~Jiang, D.~Sun, V.~Jampani, M.-H. Yang, E.~Learned-Miller, and J.~Kautz.
\newblock Super {S}lo{M}o: High quality estimation of multiple intermediate
  frames for video interpolation.
\newblock In {\em The IEEE Conference on Computer Vision and Pattern
  Recognition (CVPR)}, Jun. 2018.

\bibitem{johnson2016perceptual}
J.~Johnson, A.~Alahi, and L.~Fei-Fei.
\newblock Perceptual losses for real-time style transfer and super-resolution.
\newblock In {\em The European Conference on Computer Vision (ECCV)}, pages
  694--711. Springer, 2016.

\bibitem{kang2007motion}
S.-J. Kang, K.-R. Cho, and Y.~H. Kim.
\newblock Motion compensated frame rate up-conversion using extended bilateral
  motion estimation.
\newblock {\em IEEE Transactions on Consumer Electronics}, 53(4):1759--1767,
  2007.

\bibitem{lee2003weighted}
S.-H. Lee, O.~Kwon, and R.-H. Park.
\newblock Weighted-adaptive motion-compensated frame rate up-conversion.
\newblock {\em IEEE Transactions on Consumer Electronics}, 49(3):485--492,
  2003.

\bibitem{liu2019cyclicgen}
Y.-L. Liu, Y.-T. Liao, Y.-Y. Lin, and Y.-Y. Chuang.
\newblock Deep video frame interpolation using cyclic frame generation.
\newblock In {\em Proceedings of the 33rd Conference on Artificial Intelligence
  (AAAI)}, 2019.

\bibitem{Liu_2017_ICCV}
Z.~Liu, R.~A. Yeh, X.~Tang, Y.~Liu, and A.~Agarwala.
\newblock Video frame synthesis using deep voxel flow.
\newblock In {\em The IEEE International Conference on Computer Vision (ICCV)},
  Oct. 2017.

\bibitem{long2016learning}
G.~Long, L.~Kneip, J.~M. Alvarez, H.~Li, X.~Zhang, and Q.~Yu.
\newblock Learning image matching by simply watching video.
\newblock In {\em The European Conference on Computer Vision (ECCV)}, 2016.

\bibitem{Meyer_2018_CVPR}
S.~Meyer, A.~Djelouah, B.~McWilliams, A.~Sorkine-Hornung, M.~Gross, and
  C.~Schroers.
\newblock Phasenet for video frame interpolation.
\newblock In {\em The IEEE Conference on Computer Vision and Pattern
  Recognition (CVPR)}, Jun. 2018.

\bibitem{Meyer_2015_CVPR}
S.~Meyer, O.~Wang, H.~Zimmer, M.~Grosse, and A.~Sorkine-Hornung.
\newblock Phase-{B}ased frame interpolation for video.
\newblock In {\em The IEEE Conference on Computer Vision and Pattern
  Recognition (CVPR)}, Jun. 2015.

\bibitem{Nah_2019_CVPR_Workshops_REDS}
S.~Nah, S.~Baik, S.~Hong, G.~Moon, S.~Son, R.~Timofte, and K.~M. Lee.
\newblock {NTIRE} 2019 challenges on video deblurring and super-resolution:
  Dataset and study.
\newblock In {\em The IEEE Conference on Computer Vision and Pattern
  Recognition (CVPR) Workshops}, Jun. 2019.

\bibitem{Nah_2017_CVPR}
S.~Nah, T.~H. Kim, and K.~M. Lee.
\newblock Deep multi-scale convolutional neural network for dynamic scene
  deblurring.
\newblock In {\em The IEEE Conference on Computer Vision and Pattern
  Recognition (CVPR)}, Jul. 2017.

\bibitem{Nah_2019_CVPR_Workshops_Deblur}
S.~Nah, R.~Timofte, S.~Baik, S.~Hong, G.~Moon, S.~Son, K.~M. Lee, X.~Wang,
  K.~C. Chan, K.~Yu, C.~Dong, C.~C. Loy, Y.~Fan, J.~Yu, D.~Liu, T.~S. Huang,
  H.~Sim, M.~Kim, D.~Park, J.~Kim, S.~Y. Chun, M.~Haris, G.~Shakhnarovich,
  N.~Ukita, S.~W. Zamir, A.~Arora, S.~Khan, F.~S. Khan, L.~Shao, R.~K. Gupta,
  V.~Chudasama, H.~Patel, K.~Upla, H.~Fan, G.~Li, Y.~Zhang, X.~Li, W.~Zhang,
  Q.~He, K.~Purohit, A.~N. Rajagopalan, J.~Kim, M.~Tofighi, T.~Guo, and
  V.~Monga.
\newblock {NTIRE} 2019 challenge on video deblurring: Methods and results.
\newblock In {\em The IEEE Conference on Computer Vision and Pattern
  Recognition (CVPR) Workshops}, Jun. 2019.

\bibitem{Nah_2019_CVPR_Workshops_SR}
S.~Nah, R.~Timofte, S.~Gu, S.~Baik, S.~Hong, G.~Moon, S.~Son, K.~M. Lee,
  X.~Wang, K.~C. Chan, K.~Yu, C.~Dong, C.~C. Loy, Y.~Fan, J.~Yu, D.~Liu, T.~S.
  Huang, X.~Liu, C.~Li, D.~He, Y.~Ding, S.~Wen, F.~Porikli, R.~Kalarot,
  M.~Haris, G.~Shakhnarovich, N.~Ukita, P.~Yi, Z.~Wang, K.~Jiang, J.~Jiang,
  J.~Ma, H.~Dong, X.~Zhang, Z.~Hu, K.~Kim, D.~U. Kang, S.~Y. Chun, K.~Purohit,
  A.~N. Rajagopalan, Y.~Tian, Y.~Zhang, Y.~Fu, C.~Xu, A.~M. Tekalp, M.~A.
  Yilmaz, C.~Korkmaz, M.~Sharma, M.~Makwana, A.~Badhwar, A.~P. Singh,
  A.~Upadhyay, R.~Mukhopadhyay, A.~Shukla, D.~Khanna, A.~Mandal, S.~Chaudhury,
  S.~Miao, Y.~Zhu, and X.~Huo.
\newblock {NTIRE} 2019 challenge on video super-resolution: Methods and
  results.
\newblock In {\em The IEEE Conference on Computer Vision and Pattern
  Recognition (CVPR) Workshops}, Jun. 2019.

\bibitem{Niklaus_2018_CVPR}
S.~Niklaus and F.~Liu.
\newblock Context-aware synthesis for video frame interpolation.
\newblock In {\em The IEEE Conference on Computer Vision and Pattern
  Recognition (CVPR)}, Jun. 2018.

\bibitem{Niklaus_2017_CVPR}
S.~Niklaus, L.~Mai, and F.~Liu.
\newblock Video frame interpolation via adaptive convolution.
\newblock In {\em The IEEE Conference on Computer Vision and Pattern
  Recognition (CVPR)}, Jul. 2017.

\bibitem{Niklaus_2017_ICCV}
S.~Niklaus, L.~Mai, and F.~Liu.
\newblock Video frame interpolation via adaptive separable convolution.
\newblock In {\em The IEEE International Conference on Computer Vision (ICCV)},
  Oct. 2017.

\bibitem{Park_2019_CVPR_Workshops}
B.~Park, S.~Yu, and J.~Jeong.
\newblock Densely connected hierarchical network for image denoising.
\newblock In {\em The IEEE Conference on Computer Vision and Pattern
  Recognition (CVPR) Workshops}, Jun. 2019.

\bibitem{Peleg_2019_CVPR}
T.~Peleg, P.~Szekely, D.~Sabo, and O.~Sendik.
\newblock Im-net for high resolution video frame interpolation.
\newblock In {\em The IEEE Conference on Computer Vision and Pattern
  Recognition (CVPR)}, Jun. 2019.

\bibitem{Perazzi2016}
F.~Perazzi, J.~Pont-Tuset, B.~McWilliams, L.~{Van Gool}, M.~Gross, and
  A.~Sorkine-Hornung.
\newblock A benchmark dataset and evaluation methodology for video object
  segmentation.
\newblock In {\em The IEEE Conference on Computer Vision and Pattern
  Recognition (CVPR)}, Jun.

\bibitem{Pont-Tuset_arXiv_2017}
J.~Pont-Tuset, F.~Perazzi, S.~Caelles, P.~Arbel\'aez, A.~Sorkine-Hornung, and
  L.~{Van Gool}.
\newblock The 2017 {DAVIS} challenge on video object segmentation.
\newblock {\em arXiv preprint arXiv:1704.00675}, 2017.

\bibitem{Ranjan_2017_CVPR}
A.~Ranjan and M.~J. Black.
\newblock Optical flow estimation using a spatial pyramid network.
\newblock In {\em The IEEE Conference on Computer Vision and Pattern
  Recognition (CVPR)}, Jul. 2017.

\bibitem{space_time_super}
O.~Shahar, A.~Faktor, and M.~Irani.
\newblock Space-{T}ime super-resolution from a single video.
\newblock In {\em The IEEE Conference on Computer Vision and Pattern
  Recognition (CVPR)}, Jun. 2011.

\bibitem{Shocher_2018_CVPR}
A.~Shocher, N.~Cohen, and M.~Irani.
\newblock “zero-shot” super-resolution using deep internal learning.
\newblock In {\em The IEEE Conference on Computer Vision and Pattern
  Recognition (CVPR)}, Jun. 2018.

\bibitem{qvi_iccvw19}
L.~Siyao, X.~Xu, Z.~Pan, and W.~Sun.
\newblock Quadratic video interpolation for vtsr challenge.
\newblock In {\em The IEEE International Conference on Computer Vision (ICCV)}.

\bibitem{soomro2012ucf101}
K.~Soomro, A.~R. Zamir, and M.~Shah.
\newblock {UCF}101: A dataset of 101 human actions classes from videos in the
  wild.
\newblock {\em CRCV-TR-12-01}, Nov. 2012.

\bibitem{stich2008view}
T.~Stich, C.~Linz, G.~Albuquerque, and M.~Magnor.
\newblock View and time interpolation in image space.
\newblock In {\em Computer Graphics Forum}, volume~27, pages 1781--1787. Wiley
  Online Library, 2008.

\bibitem{Su_2019_CVPR}
H.~Su, V.~Jampani, D.~Sun, O.~Gallo, E.~Learned-Miller, and J.~Kautz.
\newblock Pixel-{A}daptive convolutional neural networks.
\newblock In {\em The IEEE Conference on Computer Vision and Pattern
  Recognition (CVPR)}, Jun. 2019.

\bibitem{Su_2017_CVPR}
S.~Su, M.~Delbracio, J.~Wang, G.~Sapiro, W.~Heidrich, and O.~Wang.
\newblock Deep video deblurring for hand-held cameras.
\newblock In {\em The IEEE Conference on Computer Vision and Pattern
  Recognition (CVPR)}, Jul. 2017.

\bibitem{Sun_2018_CVPR}
D.~Sun, X.~Yang, M.-Y. Liu, and J.~Kautz.
\newblock {PWC}-{N}et: {CNN}s for optical flow using pyramid, warping, and cost
  volume.
\newblock In {\em The IEEE Conference on Computer Vision and Pattern
  Recognition (CVPR)}, Jun. 2018.

\bibitem{Timofte_2016_CVPR}
R.~Timofte, R.~Rothe, and L.~Van~Gool.
\newblock Seven ways to improve example-based single image super resolution.
\newblock In {\em The IEEE Conference on Computer Vision and Pattern
  Recognition (CVPR)}, Jun. 2016.

\bibitem{wadhwa2013phase}
N.~Wadhwa, M.~Rubinstein, F.~Durand, and W.~T. Freeman.
\newblock Phase-based video motion processing.
\newblock {\em ACM Transactions on Graphics (TOG)}, 32(4):80, 2013.

\bibitem{qvi_nips19}
X.~Xu, L.~Siyao, W.~Sun, Q.~Yin, and M.-H. Yang.
\newblock Quadratic video interpolation.
\newblock In {\em NeurIPS}, 2019.

\bibitem{xue2019video}
T.~Xue, B.~Chen, J.~Wu, D.~Wei, and W.~T. Freeman.
\newblock Video enhancement with task-oriented flow.
\newblock {\em International Journal of Computer Vision (IJCV)},
  127(8):1106--1125, 2019.

\bibitem{Yin2019enforcing}
W.~Yin, Y.~Liu, C.~Shen, and Y.~Yan.
\newblock Enforcing geometric constraints of virtual normal for depth
  prediction.
\newblock In {\em The IEEE International Conference on Computer Vision (ICCV)},
  2019.

\bibitem{Yu_2019_CVPR_Workshops}
S.~Yu, B.~Park, and J.~Jeong.
\newblock Deep iterative down-up cnn for image denoising.
\newblock In {\em The IEEE Conference on Computer Vision and Pattern
  Recognition (CVPR) Workshops}, Jun. 2019.

\end{thebibliography}

}

\end{document}